\documentclass[letterpaper, 10 pt, conference]{ieeeconf}  

\IEEEoverridecommandlockouts                              

\usepackage{cleveref}
\usepackage{graphicx}
\usepackage{url}
\usepackage[table,xcdraw]{xcolor}
\usepackage{subcaption}

\title{\LARGE \bf
Fake or Real, Can Robots Tell? Evaluating VLM Robustness to Domain Shift in Single-View Robotic Scene Understanding}

\author{Federico Tavella$^{1}$, Amber Drinkwater$^{2}$ and Angelo Cangelosi$^{1}$
    \thanks{*This work was supported by the Centre for Robotic Autonomy in Demanding and Long-lasting Environments (CRADLE), UKRI Grant EP/X02489X/1.}
\thanks{$^{1}$Centre for Robotics and AI,
        University of Manchester, M13 9PL, Manchester, United Kingdom, \newline
        {\tt\small federico.tavella@manchester.ac.uk},
        {\tt\small angelo.cangelosi@manchester.ac.uk}}%
\thanks{$^{2}$Amentum Clean Energy Ltd, Warrington, WA3 6XF, United Kingdom,
        {\tt\small amber.drinkwater@global.amentum.com}}%
}

\begin{document}

\maketitle
\thispagestyle{empty}
\pagestyle{empty}

\begin{abstract}

Robotic scene understanding increasingly relies on Vision-Language Models (VLMs) to generate natural language descriptions of the environment. In this work, we systematically evaluate single-view object captioning for tabletop scenes captured by a robotic manipulator, introducing a controlled physical domain shift that contrasts real-world tools with geometrically similar 3D-printed counterparts that differ in texture, colour, and material. We benchmark a suite of state-of-the-art, locally deployable VLMs across multiple metrics to assess semantic alignment and factual grounding. Our results demonstrate that while VLMs describe common real-world objects effectively, performance degrades markedly on 3D-printed items despite their structurally familiar forms. We further expose critical vulnerabilities in standard evaluation metrics, showing that some fail to detect domain shifts entirely or reward fluent but factually incorrect captions. These findings highlight the limitations of deploying foundation models for embodied agents and the need for more robust architectures and evaluation protocols in physical robotic applications.

\end{abstract}

\section{Introduction}

Understanding and describing visual scenes in natural language is a fundamental capability for autonomous agents operating in human-facing environments. The emergence of powerful vision-language models (VLMs), such as BLIP \cite{li2022blip,pmlr-v202-li23q}, Flamingo \cite{alayrac2022flamingo}, and GPT-4V \cite{openai2024gpt4technicalreport}, has enabled robotic systems to leverage large pre-trained foundation models for generating image captions, recognising objects, and articulating complex scene observations in natural language. These capabilities are especially valuable in human-robot collaboration, where robots must identify and describe task-relevant objects within dynamic, cluttered environments.

\begin{figure}[ht]
    \centering
    \includegraphics[width=.45\textwidth]{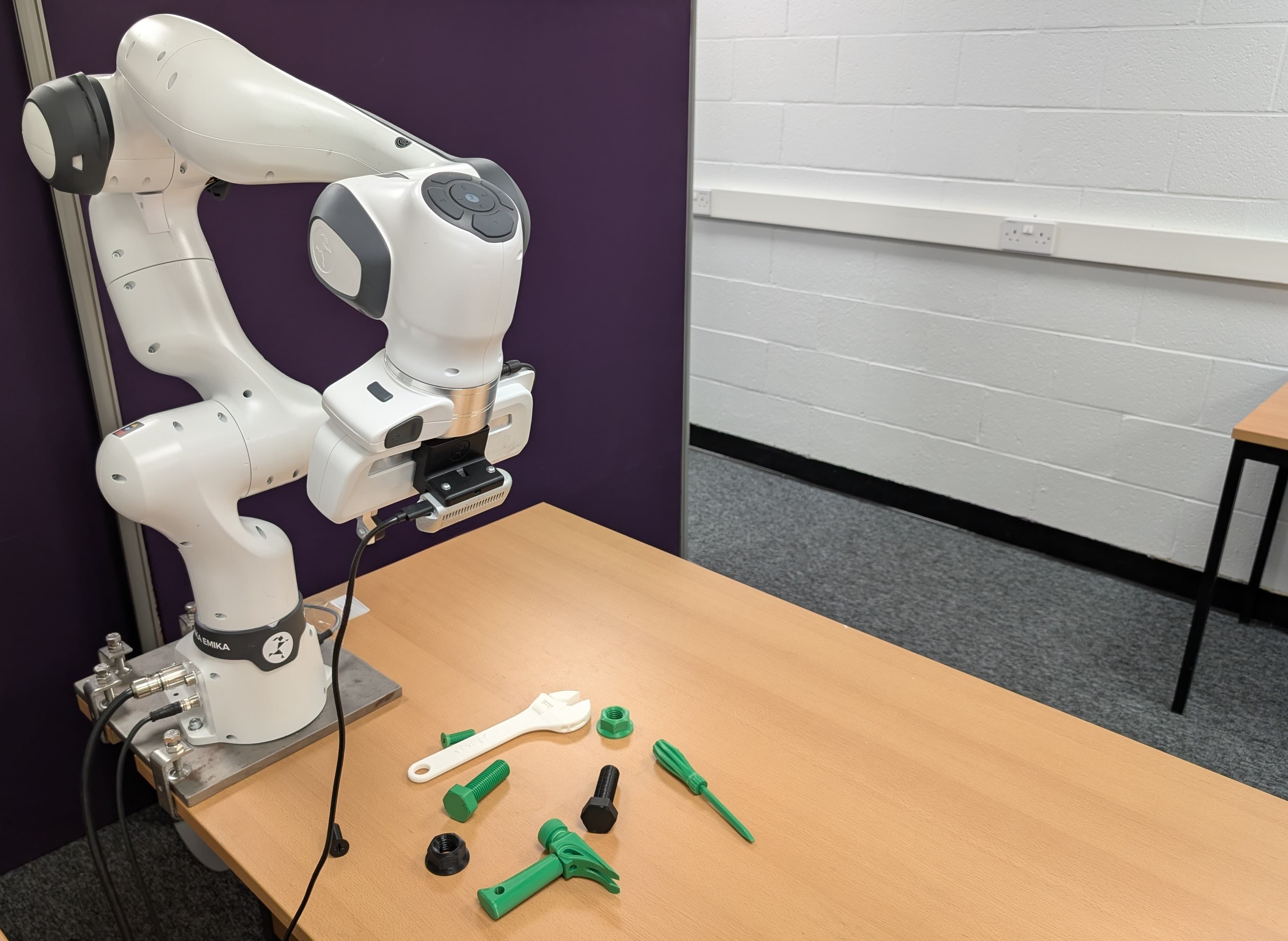}
    \caption{Experimental setup. A Franka Emika Research 3 robotic manipulator equipped with a gripper-mounted RGB camera captures single overhead images of tabletop scenes for single-view VLM evaluation.}
    \label{fig:enter-label}
\end{figure}

Despite these advances, significant challenges remain. Most captioning systems operate on single-view images, which are frequently compromised by occlusions, ambiguous viewpoints, or partial visibility -- conditions that are routine in real-world deployments. For embodied agents such as robotic arms, initial scene understanding typically depends on a single observation (e.g., a top-down inspection pose) prior to grasping or manipulation. The robustness of VLMs to out-of-distribution (OOD) objects under these single-view constraints therefore constitutes a critical bottleneck. Two questions are central: How well do current VLMs withstand physical domain shifts in such scenarios? And how can this robustness be reliably quantified?
In this paper, we systematically evaluate visual-linguistic captioning strategies for tabletop object understanding under single-view inputs exclusively. \Cref{fig:enter-label} illustrates our experimental setup. Our investigation centres on two aspects: (i) the choice of captioning model, assessed across a suite of state-of-the-art, locally deployable VLMs; and (ii) the metrics and criteria used to evaluate caption quality and factual grounding. A distinguishing feature of our study is a real-world generalisation test in which models are evaluated not only on genuine objects but also on geometrically similar 3D-printed counterparts that differ in texture, colour, and material properties. This controlled contrast allows us to examine whether current VLMs can accurately recognise and describe physical objects whose surface appearance deviates substantially from training-distribution expectations, an important step towards robust physical-world understanding.
Our findings reveal that while VLMs can caption common objects effectively from a single viewpoint, performance degrades substantially when describing 3D-printed objects with altered textures and colours. We additionally validate and compare a range of metrics drawn from natural language processing and computer vision to measure caption similarity, exposing critical vulnerabilities that arise when these metrics are applied in the context of physical domain shifts.

\begin{figure*}[ht]
    \centering
    \includegraphics[width=.99\textwidth]{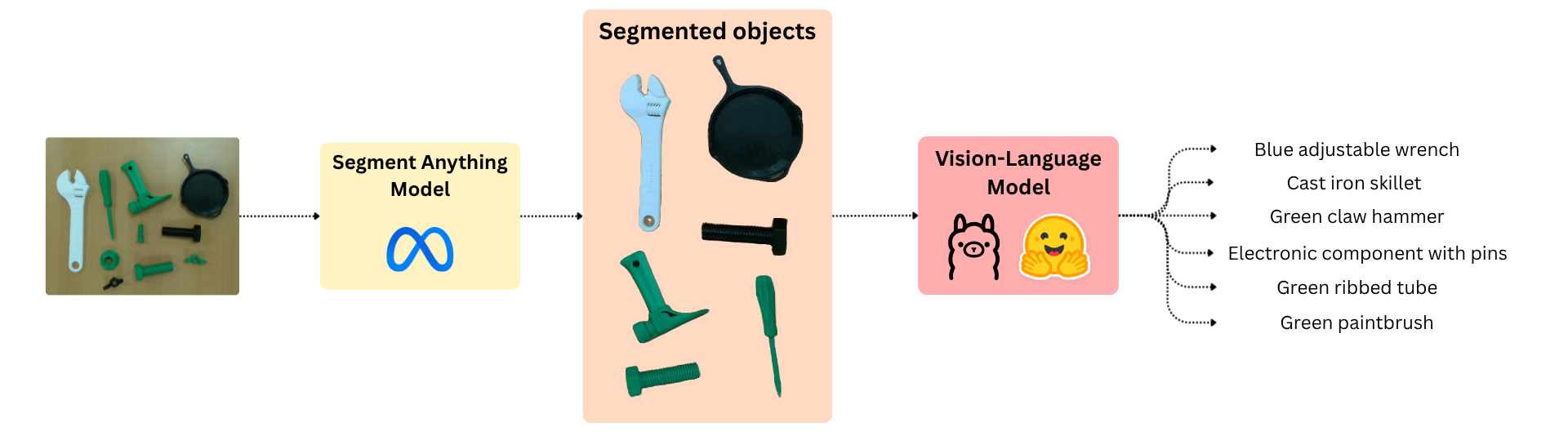}
    \caption{Our evaluation pipeline for object captioning. The Segment Anything Model isolates objects from a scene, which are then described by a Vision-Language Model . The example outputs illustrate the variance in performance, including successful recognitions (e.g., ``Cast iron skillet'') and common failure modes where the VLM misidentifies an object (e.g., mistaking a nut for an ``Electronic component'').}
    \label{fig:diagram}
\end{figure*}

\section{Related Work}

\subsection{Vision-Language Models for Captioning}

Image captioning has progressed substantially over the past decade. Early systems such as Show and Tell \cite{vinyals2015show} and Show, Attend and Tell \cite{xu2015show} established the CNN-RNN paradigm, coupling visual feature extraction with sequential language generation. The subsequent shift to transformer-based architectures brought significant gains: BLIP \cite{li2022blip,pmlr-v202-li23q} and Flamingo \cite{alayrac2022flamingo} leverage large-scale vision-language pre-training to achieve strong generalisation across diverse benchmarks, while CLIP \cite{radford2021learning_clip} introduced contrastive image-text alignment as a foundation for zero-shot recognition, influencing both model design and evaluation methodology. More recently, multimodal large language models such as GPT-4V \cite{openai2024gpt4technicalreport} have extended these capabilities to complex visual reasoning and naturalistic description in few-shot and zero-shot settings.

\subsection{VLMs in Robotics}
There is growing interest in repurposing large VLMs as general-purpose perception backbones for robotic systems \cite{rt2, alayrac2022flamingo}. Models such as PaLI-X \cite{chen2023pali} and OpenFlamingo \cite{awadalla2023openflamingo} have been adapted to produce rich semantic embeddings from raw images and language goals, enabling open-vocabulary recognition and web-scale knowledge transfer to robot control \cite{rt2, li2023vision}. Language grounding allows agents to interpret instructions, describe their surroundings, and communicate intent -- capabilities assessed in benchmarks such as ALFRED \cite{shridhar2020alfred} -- while systems such as SayCan \cite{ahn2022icanisay} and LLM-Grounder \cite{yang2024llm_grounder} connect natural language to physical affordances and object references respectively. More broadly, VLMs are increasingly deployed as semantic planners that decompose natural-language commands into skill sequences and grasp targets \cite{liu2025robodexvlm, zhao2024vlmpc}, making the robustness of the underlying VLM a safety-critical property of the overall system.

\subsection{Out-of-Distribution Perception and Domain Shift}
Despite the remarkable capabilities of pre-trained VLMs, their behaviour under out-of-distribution (OOD) conditions remains an open research problem. Domain generalisation methods seek to improve the stability of vision-language alignment under unseen distributions \cite{zhu2024vision, chen2024practicaldg, pmlr-v235-huang24o}, while dedicated benchmarks have demonstrated that large VLMs remain highly susceptible to both semantic and covariate shifts \cite{noda2025benchmark, miyai2024generalized2}. In robotics, methods such as RobustVLA \cite{zhang2025robustvla} and RETAIN \cite{yadav2026robust} propose robustness-aware fine-tuning schemes to improve VLA policy reliability under observation noise, actuation disturbances, and scene variation. However, this literature focuses predominantly on digital perturbations or sim-to-real transfer; robustness to physical, material-based domain shifts remains largely unexplored.

\subsection{Summary}

Our work bridges these fields through a focused comparison of single-view captioning pipelines for tabletop robotic perception. Motivated by the active deployment of VLMs as semantic planners in embodied agents \cite{liu2025robodexvlm, zhao2024vlmpc}, we evaluate a suite of open-source models under a controlled physical domain shift -- contrasting performance on real objects against geometrically similar but materially distinct 3D-printed counterparts -- and examine both model performance and the reliability of the metrics used to assess it, exposing vulnerabilities in applying foundation models to physical-world perception.

\section{Methodology}

This work systematically evaluates the robustness of open-source VLMs under a controlled physical domain shift. We benchmark a suite of state-of-the-art locally deployable models on a single-view object captioning task, comparing their ability to describe two object sets: real-world tools and geometrically similar but materially distinct 3D-printed counterparts. An overview of our approach is illustrated in \Cref{fig:diagram}.

\subsection{Data}

\begin{figure}[ht]
    \centering
    \begin{subfigure}[b]{0.48\textwidth}
        \centering
        \includegraphics[width=\textwidth]{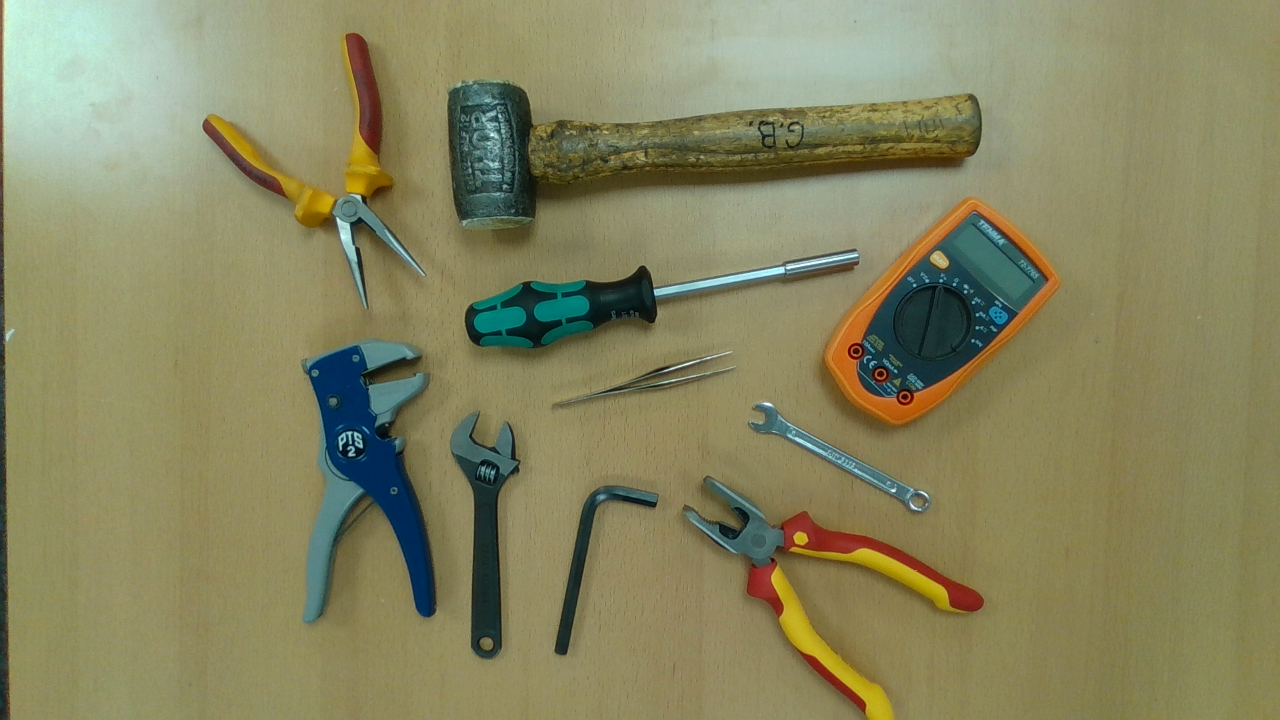}
        \caption{Real objects}
        \label{fig:objs_real}
    \end{subfigure}
    \vfill
    \begin{subfigure}[b]{0.48\textwidth}
        \centering
        \includegraphics[width=\textwidth]{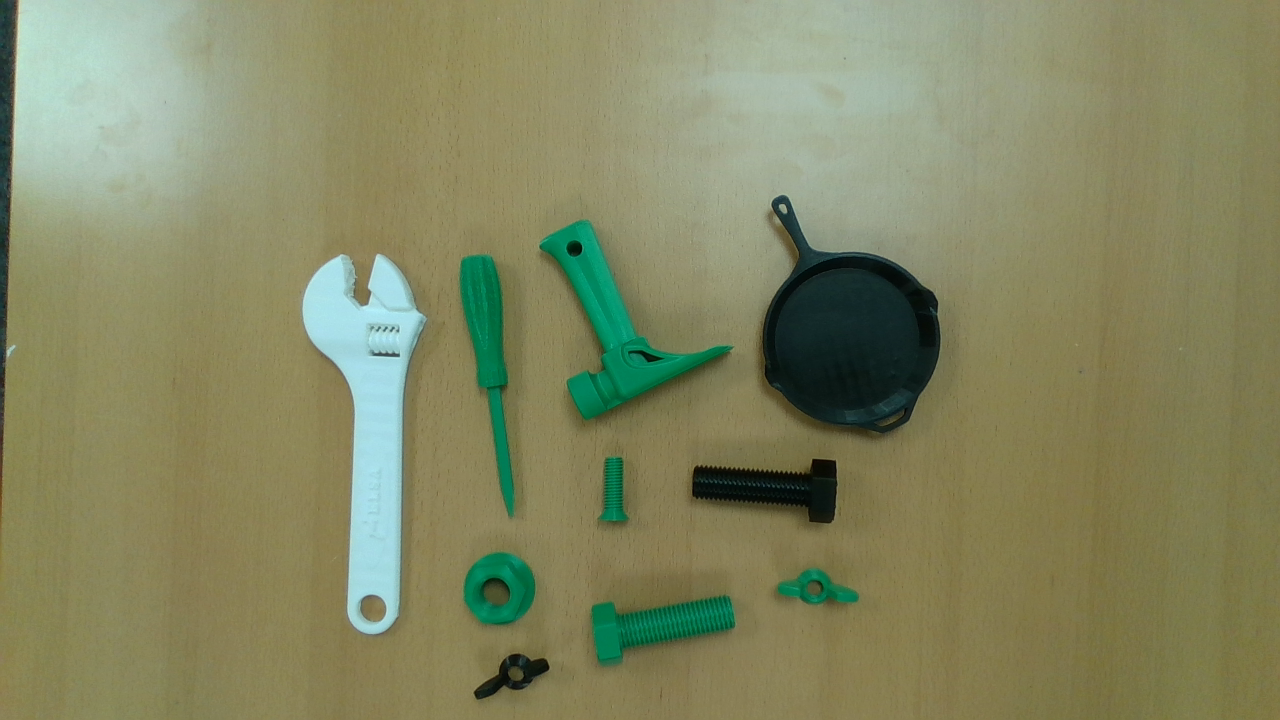}
        \caption{3D printed objects}
        \label{fig:objs_3d}
    \end{subfigure}
    \caption{Our test objects. Each set is composed by 10 different objects.}
    \label{fig:objects}
\end{figure}

Our experiments employ two object sets to construct a controlled physical domain shift. The first comprises real-world tools made from authentic materials such as metal and wood; the second consists of geometrically similar objects fabricated from 3D-printed plastic. While the 3D-printed items broadly preserve the expected shapes of their real-world counterparts, they differ substantially in texture and material properties, making them an effective proxy for out-of-distribution (OOD) data with which to probe model robustness. This design directly tests whether VLMs, typically trained on photorealistic web imagery, can generalise beyond the surface-level appearance cues on which they are known to rely. \Cref{tab:objects} lists all objects in both sets, and \Cref{fig:objects} illustrates representative examples.

\begin{table}[ht]
\centering
\resizebox{.49\textwidth}{!}{%
\begin{tabular}{cc}
\textbf{Real-world objects}                                                                             & \textbf{3D printed objects}                                                               \\ \hline \hline
orange and black digital multimeter                                                       & plastic white adjustable wrench                                                   \\ \hline
grey and blue wire stripper                                                               & plastic black cast iron pan                                                       \\ \hline
metal combination wrench                                                                  & plastic green hammer                                                              \\ \hline
metal tweezers                                                                            & plastic black hexagonal bolt                                                      \\ \hline
\begin{tabular}[c]{@{}c@{}}screwdriver with \\ teal and black handle\end{tabular}         & plastic green hexagonal bolt                                                      \\ \hline
Allen hexagonal wrench                                                                    & \begin{tabular}[c]{@{}c@{}}plastic green hexagonal nut\\ with flange\end{tabular} \\ \hline
Adjustable wrench                                                                         & plastic black wing nut                                                           \\ \hline
\begin{tabular}[c]{@{}c@{}}Combination pliers \\ with red and yellow handles\end{tabular} & plastic green wing nut                                                           \\ \hline
\begin{tabular}[c]{@{}c@{}}Needle-nose pliers \\ with red and yellow handles\end{tabular} & plastic green nut                                                                \\ \hline
hammer with wooden handle                                                                 & plastic green screwdriver                                                         \\ \hline \hline
\end{tabular}%
}
\caption{Description/caption of each of the 10 objects for the two sets.}
\label{tab:objects}
\end{table}

\subsection{Segmentation}

To isolate individual objects for captioning, we employ the Segment Anything Model 2 (SAM2) \cite{ravi2024sam} for its zero-shot segmentation capabilities, which enable robust object-level parsing without requiring fine-tuning or manual annotation. We apply SAM2 in automatic mode, in which the model generates dense segmentation masks over the full scene without user-defined prompts. The resulting masks are then filtered programmatically to remove small spurious regions and any mask fully contained within another, ensuring that each retained mask corresponds to a distinct, well-defined object.

\subsection{Captioning (VLMs)}

To generate natural language descriptions of the segmented objects, we evaluate eight state-of-the-art open-source VLMs: SmolVLM2 \cite{SmolVLM222BInstructHugging}, BLIP-2 \cite{pmlr-v202-li23q}, Gemma 3n E4B \cite{googleGemma3nModel}, LLaMA 3.2 \cite{grattafiori2024llama3herdmodels}, Mistral Small 3.1 \cite{mistralMistralSmall}, Gemma 3 \cite{team2025gemma}, Qwen 2.5 VL \cite{bai2025qwen25vltechnicalreport}, and LLaVA 1.6 \cite{liu2023visual, liu2024improved}. All models are open-source and locally deployable on consumer-grade hardware, with parameter counts not exceeding 35 billion, a constraint that ensures full reproducibility and provides a practical benchmark for researchers without access to proprietary APIs or large-scale compute. BLIP-2 is loaded via HuggingFace\footnote{\url{https://huggingface.co}}; all remaining models are served through Ollama\footnote{\url{https://ollama.com}}. Model selection was further bounded by available hardware: all models were verified to run on a single NVIDIA RTX 4090 (24 GB VRAM).
All models receive an identical prompt:
\textit{``Describe in detail the object in the image in a few words. Maintain a concise, factual tone. Ignore the background. Here is an example of your output if you are given an image of a red apple on a table: red apple.''}
This standardised prompt ensures that observed performance differences are attributable to model capability rather than prompt engineering, and the inclusion of a concrete output example helps suppress verbose or conversational responses that would confound metric-based evaluation.

\subsection{Metrics}

We employ multiple evaluation metrics to assess the quality of the generated captions. CLIPScore \cite{hessel-etal-2021-clipscore} (based on the \texttt{ViT-B/32} model) is used to measure the semantic similarity between the generated caption and the image itself, leveraging the joint embedding space of the CLIP model \cite{radford2021learning_clip}. For text-to-text comparison against the ground truth, we use a focused set of four complementary metrics. We use ROUGE \cite{lin2004rouge} to measure n-gram recall, determining if essential words are captured. To evaluate alignment with human consensus, we use CIDEr \cite{Vedantam2015cider}, a standard for image captioning tasks. For a deeper semantic understanding, we use BERTScore \cite{zhang2019bertscore}, which compares contextual embeddings to capture meaning beyond simple word overlap; in our implementation, we use \texttt{deberta-xlarge-mnli}\footnote{\url{https://huggingface.co/microsoft/deberta-xlarge-mnli}} to calculate the BERTScore. Finally, to incorporate a state-of-the-art evaluation, we use GPTScore \cite{fu2024gptscore}, which leverages a large language model to score a caption based on its likelihood given the ground truth as context. While our primary analysis relies on ROUGE, CIDEr, BERTScore, and GPTScore, we include CLIPScore to critically evaluate its effectiveness in this domain-shift context.
By combining multiple metrics, we get a more complete view of model performance (balancing lexical accuracy, semantic relevance, fluency, and diversity) and reduce the risk of overfitting to the characteristics of any one metric. This approach is particularly important when evaluating open-ended tasks like captioning or text generation, where there are many valid outputs. It is worth noting that CLIPScore, ROUGE and BERTScore metrics range from 0 to 1, CIDEr from 0 to 10, and GPTScore has values below 0.

\subsection{Experimental procedure}

We evaluate the ability of VLMs to recognise objects from a top-down perspective. To this end, the robot arm positions the camera directly above each object, capturing a single overhead image. We use SAM2 to segment the scene and generate an individual mask for each object. These segmented and cropped object images are then passed to the VLM, which produce a single caption per object. The generated captions are subsequently compared against both the visual input and human-provided descriptions to assess the accuracy and alignment of the model’s output with human perception. We repeat this process for both sets of objects.

\section{Results}

This section presents our single-view evaluation, first detailing the quantitative performance of the models across our two object sets, followed by qualitative examples that illustrate key findings.

\begin{table*}[t]
\centering
\resizebox{\textwidth}{!}{%
\begin{tabular}{l|cccccccccc}
\textbf{}         & \multicolumn{2}{c}{\textbf{CLIPScore}}                              & \multicolumn{2}{c}{\textbf{ROUGE}}                                  & \multicolumn{2}{c}{\textbf{CIDEr}}                                  & \multicolumn{2}{c}{\textbf{BERTScore}}                              & \multicolumn{2}{c}{\textbf{GPTScore}}                                 \\
\textbf{VLM}      & \cellcolor[HTML]{EFEFEF}\textbf{Real}        & \textbf{3D}          & \cellcolor[HTML]{EFEFEF}\textbf{Real}        & \textbf{3D}          & \cellcolor[HTML]{EFEFEF}\textbf{Real}        & \textbf{3D}          & \cellcolor[HTML]{EFEFEF}\textbf{Real}        & \textbf{3D}          & \cellcolor[HTML]{EFEFEF}\textbf{Real}         & \textbf{3D}           \\ \hline \hline
SmolVLM2 (2.2B)   & \cellcolor[HTML]{EFEFEF}0.29 ± 0.04          & 0.28 ± 0.05          & \cellcolor[HTML]{EFEFEF}0.34 ± 0.27          & \textbf{0.45 ± 0.15} & \cellcolor[HTML]{EFEFEF}1.19 ± 1.16          & \textbf{0.90 ± 0.80} & \cellcolor[HTML]{EFEFEF}0.30 ± 0.20          & 0.28 ± 0.20          & \cellcolor[HTML]{EFEFEF}-8.69 ± 2.72          & -8.35 ± 1.60          \\
BLIP2 (2.7B)      & \cellcolor[HTML]{EFEFEF}\textbf{0.32 ± 0.05} & 0.31 ± 0.04          & \cellcolor[HTML]{EFEFEF}0.34 ± 0.22          & 0.20 ± 0.06          & \cellcolor[HTML]{EFEFEF}0.94 ± 0.81          & 0.13 ± 0.17          & \cellcolor[HTML]{EFEFEF}0.42 ± 0.23          & 0.35 ± 0.16          & \cellcolor[HTML]{EFEFEF}\textbf{-3.53 ± 0.71} & -4.44 ± 0.73          \\
Gemma 3n (4B)     & \cellcolor[HTML]{EFEFEF}0.29 ± 0.05          & 0.30 ± 0.03          & \cellcolor[HTML]{EFEFEF}0.32 ± 0.33          & 0.35 ± 0.20          & \cellcolor[HTML]{EFEFEF}1.24 ± 1.59          & 0.94 ± 1.30          & \cellcolor[HTML]{EFEFEF}0.35 ± 0.33          & \textbf{0.42 ± 0.16} & \cellcolor[HTML]{EFEFEF}-5.23 ± 2.01          & -5.74 ± 1.61          \\
Llama 3.2 (11B)   & \cellcolor[HTML]{EFEFEF}0.29 ± 0.04          & 0.29 ± 0.02          & \cellcolor[HTML]{EFEFEF}0.20 ± 0.15          & 0.19 ± 0.09          & \cellcolor[HTML]{EFEFEF}0.30 ± 0.41          & 0.18 ± 0.25          & \cellcolor[HTML]{EFEFEF}0.24 ± 0.19          & 0.28 ± 0.13          & \cellcolor[HTML]{EFEFEF}-3.73 ± 0.84          & -4.55 ± 1.43          \\
Mistral 3.1 (24B) & \cellcolor[HTML]{EFEFEF}0.31 ± 0.04          & \textbf{0.31 ± 0.03} & \cellcolor[HTML]{EFEFEF}0.33 ± 0.20          & 0.31 ± 0.18          & \cellcolor[HTML]{EFEFEF}0.98 ± 0.99          & 0.51 ± 0.81          & \cellcolor[HTML]{EFEFEF}0.35 ± 0.16          & 0.30 ± 0.14          & \cellcolor[HTML]{EFEFEF}-5.46 ± 2.02          & -5.77 ± 1.44          \\
Gemma 3 (27B)     & \cellcolor[HTML]{EFEFEF}0.29 ± 0.04          & 0.30 ± 0.03          & \cellcolor[HTML]{EFEFEF}0.32 ± 0.27          & 0.35 ± 0.18          & \cellcolor[HTML]{EFEFEF}1.26 ± 1.48          & 0.80 ± 1.01          & \cellcolor[HTML]{EFEFEF}0.41 ± 0.26          & 0.35 ± 0.15          & \cellcolor[HTML]{EFEFEF}-5.46 ± 1.31          & -6.34 ± 0.85          \\
Qwen 2.5 VL (32B) & \cellcolor[HTML]{EFEFEF}0.31 ± 0.03          & \textbf{0.31 ± 0.03} & \cellcolor[HTML]{EFEFEF}\textbf{0.42 ± 0.32} & 0.29 ± 0.14          & \cellcolor[HTML]{EFEFEF}\textbf{1.43 ± 1.83} & 0.60 ± 0.71          & \cellcolor[HTML]{EFEFEF}\textbf{0.52 ± 0.29} & 0.31 ± 0.12          & \cellcolor[HTML]{EFEFEF}-4.25 ± 0.82          & \textbf{-4.23 ± 0.37} \\
LLaVA 1.6 (34B)   & \cellcolor[HTML]{EFEFEF}0.30 ± 0.03          & 0.30 ± 0.03          & \cellcolor[HTML]{EFEFEF}0.16 ± 0.10          & 0.22 ± 0.16          & \cellcolor[HTML]{EFEFEF}0.32 ± 0.41          & 0.16 ± 0.32          & \cellcolor[HTML]{EFEFEF}0.25 ± 0.20          & 0.22 ± 0.10          & \cellcolor[HTML]{EFEFEF}-3.79 ± 0.77          & -5.52 ± 1.33          \\ \hline \hline
\end{tabular}%
}
\caption{Quantitative results for the single-view object captioning task. Performance is compared across eight open-source VLMs on the Real-world and 3D-printed object sets. All scores are reported as mean and standard deviation across the 10 objects in each set. For CLIPScore, ROUGE, CIDEr, and BERTScore, higher is better. For GPTScore, scores closer to zero indicate better performance. The top-performing model for each VLM and object set over a certain metric is highlighted in bold.}
\label{tab:results}
\end{table*}
\subsection{Quantitative Analysis}The results of our single-view evaluation, presented in \Cref{tab:results}, reveal several key insights into the performance and robustness of the evaluated open-source VLMs.The most significant finding is a consistent, quantifiable degradation in performance across nearly all models when captioning 3D-printed objects compared to their real-world counterparts. This confirms our primary hypothesis that the domain shift introduced by altered textures and material properties poses a substantial challenge to current VLMs. The effect is most pronounced in the CIDEr metric, which is designed to measure consensus with human-annotated descriptions. Qwen 2.5 VL, the top-performing model overall, sees its CIDEr score fall by over 50\% -- from an average 1.43 on real objects to 0.60 on the 3D-printed set. Comparable drops are observed across BLIP2 (0.94 to 0.13) and Gemma 3 (1.26 to 0.80), confirming that this vulnerability is a systemic property of the model class rather than an architectural anomaly.The results also establish a clear performance hierarchy. Qwen 2.5 VL (32B) is the standout model, achieving the highest ROUGE and CIDEr scores on real objects while maintaining competitive performance on the more challenging 3D-printed set -- suggesting that its architecture and training confer a meaningful advantage for this robotic captioning task. At the other end of the spectrum, Llama 3.2 (11B) and LLaVA 1.6 (34B) consistently rank lower, particularly on CIDEr, indicating that their captions diverge more substantially from human-annotated ground truth. Notably, SmolVLM2 (2.2B) achieves highly competitive ROUGE performance on the 3D-printed set (0.45), outperforming both Llama 3.2 and LLaVA 1.6 despite being a fraction of their size. This suggests that parameter count alone does not determine robustness to physical domain shifts. 

\subsection{Qualitative Examples}
A qualitative review of the generated captions complements the quantitative results, providing an intuitive illustration of both model capabilities and the severity of the physical domain shift.
For real-world objects, top-performing models produced accurate, concise descriptions that were appropriately rewarded by the evaluation metrics. When presented with the ground truth label "metal tweezers," for instance, Qwen 2.5 VL generated the exact caption "metal tweezers," achieving the maximum CIDEr score of 5.00. Similarly, SmolVLM2 described the ground truth "orange and black digital multimeter" as "orange and black multimeter," earning a strong CIDEr score of 4.35.
This perceptual reliability broke down substantially on the 3D-printed set, where model outputs frequently exhibited catastrophic semantic drift. In many cases, models failed to identify the object's core functional identity. When presented with a "plastic black hexagonal bolt," Gemma 3 produced the caption "Electronic component with pins," receiving a CIDEr score of 0.0. In more extreme instances, object identity was hallucinated entirely: faced with a "plastic black wing nut," Mistral 3.1 generated the description "Dark brown oval-shaped chocolate with a light brown filling," also scoring 0.0. Captions of this kind would cause an embodied agent to fundamentally mischaracterise the object, with direct consequences for safe and appropriate grasping strategies.
Our qualitative analysis also exposed a critical weakness in GPTScore. Despite its status as a state-of-the-art LLM-based metric, it occasionally failed to penalise factually incorrect captions, appearing to reward descriptive fluency over semantic accuracy. When LLaVA 1.6 described a "plastic black hexagonal bolt" as "a firearm with a dark finish and a visible barrel, grip" -- a misidentification with serious safety implications in a real deployment -- the caption received a GPTScore of -3.86, among the least-penalised scores in our evaluation. This suggests that the underlying language model was misled by the grammatical coherence and descriptive detail of a semantically incorrect response.

\subsection{Analysis of Evaluation Metrics}
Beyond model performance, our study offers a critical assessment of the evaluation metrics themselves when applied to physical domain shift in robotic scene understanding.
CLIPScore proved entirely unsuitable for this task. As shown in \Cref{tab:results}, its values exhibited negligible variation between the real and 3D-printed object sets, demonstrating a fundamental insensitivity to the domain shift under investigation. It was therefore excluded from the primary comparative analysis.
The n-gram-based metrics, ROUGE and CIDEr, were more effective at capturing the principal performance trend: a clear degradation on the 3D-printed set. CIDEr was particularly sensitive to semantic drift -- its scores for Qwen 2.5 VL fell by over 50\% (from 1.43 to 0.60) -- reflecting its design emphasis on penalising captions that diverge from human consensus descriptions. However, both metrics share a structural limitation: they operate at the level of lexical overlap and are therefore liable to penalise valid paraphrases while remaining blind to sentence-level quality and grammatical coherence.
BERTScore addresses this limitation by operating in a semantic embedding space, making it better suited to recognising paraphrase and capturing meaning beyond exact wording. In our evaluation, however, it proved less discriminative for this physical task. Scores across all models were compressed into a narrow range (0.22–0.52), making it difficult to distinguish top performers from average ones. While BERTScore reliably captured broad semantic correctness, it lacked the granularity necessary to reflect the quality distinctions apparent to a human reader.
GPTScore, as an LLM-based metric, aims to provide a more holistic assessment of caption quality. However, our findings reveal a significant vulnerability when it is applied to safety-critical physical systems. The metric exhibited high variance across objects and, most concerning for autonomous agent deployment, failed to penalise factually incorrect captions. As noted above, LLaVA 1.6's misidentification of a hexagonal bolt as a firearm received a relatively high GPTScore of -3.86, suggesting the evaluator was misled by descriptive fluency and grammatical well-formedness. GPTScore captures a form of linguistic quality, but it cannot be relied upon to enforce the factual grounding that safe and accurate robotic manipulation demands.

\section{Discussion}

This work presents a systematic evaluation of open-source Vision-Language Models deployed for single-view robotic scene understanding. Through a controlled physical domain shift, we quantified the performance degradation that arises when VLMs are presented with objects that preserve familiar geometry but deviate from training-distribution appearance in texture and material. Our findings reveal a consistent and significant domain sensitivity across all evaluated models: while most VLMs captioned real-world tools effectively, their accuracy degraded substantially on the 3D-printed set. This confirms that current models rely heavily on surface-level features -- such as specific textures and material properties -- encoded during web-scale pre-training, and struggle to generalise beyond them. In the context of embodied AI, this is a critical limitation. An autonomous agent that cannot accurately identify an out-of-distribution object will propagate that error directly into downstream grasping, manipulation, and task planning.

Our metric analysis yields equally important practical conclusions. CLIPScore proved entirely unsuitable for this evaluation, exhibiting a fundamental insensitivity to the physical domain shift. More broadly, our results expose the trade-offs inherent in different metric classes. N-gram-based metrics such as ROUGE and CIDEr effectively captured the primary performance degradation but remain blind to semantic nuance and can penalise valid paraphrases. BERTScore, which operates over semantic embeddings and is better suited to capturing meaning beyond lexical overlap, was less discriminative in this task, compressing scores across models into a narrow range. GPTScore, despite its sophistication as an LLM-based evaluator, proved susceptible to descriptive fluency, at times failing to penalise factually incorrect captions. In robotic applications, such failures are consequential: an evaluation metric that rewards a well-formed but semantically incorrect description provides false confidence to a system that may then execute an unsafe or inappropriate action. Reliable assessment for physical deployment therefore requires a complementary suite of metrics rather than reliance on any single measure.

Several directions for future work follow naturally from these findings. The experiments reported here were conducted on a dataset of ten real-world and ten 3D-printed objects; validation on a larger and more varied object set is necessary to confirm that these conclusions generalise across diverse domestic and industrial environments. Equally, while this study focused on material and texture shifts, future work should examine other sources of variation inherent to dynamic robotic workspaces, including changes in illumination, viewpoint, and occlusion.
Addressing the observed performance degradation is, however, the most pressing avenue for future investigation. We identify three technical directions that merit exploration:

\begin{itemize}
\item \textbf{Multimodal Sensory Fusion:} Integrating depth sensing (RGB-D) or tactile feedback could supply geometric and physical priors that compensate for visually misleading surface appearances, anchoring model predictions in physical reality rather than appearance alone.
\item \textbf{In-Context Learning and Test-Time Adaptation:} Providing VLMs with few-shot prompts containing examples of domain-shifted objects (such as textureless or 3D-printed items) could dynamically recalibrate semantic mappings without requiring full retraining, offering a lightweight path to improved robustness.
\item \textbf{Targeted Domain Randomisation:} Fine-tuning on datasets intentionally augmented with synthetic, 3D-printed, or materially diverse objects could improve zero-shot robustness to material and texture shifts, better equipping models for deployment in open-world physical environments.
\end{itemize}

Taken together, these directions point towards a broader agenda: embedding physical-world robustness as a first-class requirement in the design, training, and evaluation of VLMs intended for embodied deployment.



\bibliographystyle{IEEEbib}
\bibliography{refs,refs-ral}

\end{document}